\documentclass{article}
\usepackage{spconf, amsmath, amssymb, graphicx}
\usepackage{paralist, url}
\usepackage{algpseudocode}
\usepackage{algorithm}
\usepackage{amsmath,epsfig, bm}

\ninept

\title{Toward Computation and Memory Efficient Neural Network Acoustic Models with Binary Weights and Activations}
\name{Liang Lu}
\address{Toyota Technological Institute at Chicago \\
\texttt{\small llu@ttic.edu}%
}

\begin{document}
\maketitle

\begin{abstract}
Neural network acoustic models have significantly advanced state of the art speech recognition over the past few years. However, they are usually computationally expensive due to the large number of matrix-vector multiplications and nonlinearity operations. Neural network models also require significant amounts of memory for inference because of the large model size. For these two reasons, it is challenging to deploy neural network based speech recognizers on resource-constrained platforms such as embedded devices. This paper investigates the use of binary weights and activations for computation and memory efficient neural network acoustic models. Compared to real-valued weight matrices, binary weights require much fewer bits for storage, thereby cutting down the memory footprint. Furthermore, with binary weights or activations, the matrix-vector multiplications are turned into addition and subtraction operations, which are computationally much faster and more energy efficient for hardware platforms. In this paper, we study the applications of binary weights and activations for neural network acoustic modeling, reporting encouraging results on the WSJ and AMI corpora. 
\end{abstract}
\begin{keywords}
Neural networks, Binary weights, Binary activations, Speech recognition
\end{keywords}

\section{Introduction}

Neural networks have been shown to be extremely powerful in a wide range of machine learning tasks, evidenced by recent significant progress in tasks such as speech recognition~\cite{hinton2012deep, saon2017english}, machine translation~\cite{bahdanau2014neural, wu2016google} and image recognition~\cite{krizhevsky2012imagenet}. However, computations in neural networks are usually much more expensive than prior approaches, as they involve a large number of matrix-vector multiplications followed by nonlinear activation functions.  Furthermore, model training and inference with neural networks also require a significant amount of memory due to the large size of the mode, as in nowadays neural network models, the number of hidden units can be thousands for each layer. As a result, neural network models are usually trained on GPUs with significant speedups via parallelization, and the models are usually deployed in the cloud for inference to address the memory issue.

Computation and memory efficient neural networks have been an appealing research topic for both deep learning and application researchers, as they enable local deep learning applications such as deploying speech and image recognition on embedded devices without access to the cloud. This problem has been addressed by many researchers in different ways. A large fraction of the prior works aims at training a small mode -- in terms of the number of model parameters -- that can approach the accuracy of a larger model. In this work, we are inspired by~\cite{courbariaux2015binaryconnect, courbariaux2016binarized} to investigate using binary weights and activations to replace the real-valued weights and activations in neural networks. The motivation is that compared to real-valued weights, binary weights require significantly fewer bits for storage, thereby cutting down the memory footprint. From a computational perspective, binary weighs or activations turn the matrix-vector multiplications into additions and subtractions, which are much faster and more energy efficient for hardware. With both binary weights and activations, the computation will be even simpler and faster as the multiplications become only {\tt XOR} operations, which can be implemented very efficiently on hardware. 

Compared to the pilot study of this idea for image classification on relative small datasets~\cite{courbariaux2015binaryconnect, courbariaux2016binarized} (i.e., MNIST, CIFAR-10 and SVHN), in this paper, we investigate neural networks with binary weights and activations in the context of large vocabulary speech recognition. In particular, we focus on feedforward neural networks as they are simpler in terms of training algorithms and are computationally cheaper for fast turnaround for experiments. Our training algorithms are slightly differently from~\cite{courbariaux2015binaryconnect, courbariaux2016binarized} due to the difference in our model and the task itself, which are detailed in Section \ref{sec:bnn}. Our study is mainly based on the WSJ1 corpus with some additional experiment carried out on the AMI database. 

\subsection{Related Work}
\label{sec:prior}

In both speech recognition and deep learning in general, there have been a number of approaches for small memory footprint and computation efficient neural networks. One approach is {\it teacher-student} training, also known as model compression~\cite{bucilu2006model} or knowledge distillation~\cite{hinton2015distilling}, where a large and computationally expensive teacher model (or an ensemble of models) is used to predict the soft targets for training the smaller student model. As discussed in~\cite{hinton2015distilling}, the soft targets provided by the teacher encode the generalization power of the teacher model, and the student model trained using these labels is observed to perform better than the same model trained with hard labels~\cite{ba2014deep, romero15_fitnet}. Some successful examples of using this approach for speech recognition are~\cite{li2014learning, lu2016knowledge, wong2016sequence, cui2017knowledge}. 

Motivated by the argument that neural networks with dense connections are over-parameterized, another branch of works is to replace the full-rank linear matrices in neural networks by products of low-rank structured matrices. Particular examples include the Toeplitz-like structured transforms studied in~\cite{sindhwani2015structured}, and the discrete cosine transform (DCT) used in~\cite{moczulski2015acdc} to approximate the weight matrices in neural networks. With those structured transforms, the number of trainable parameters is significantly smaller, thereby reducing the amount of memory required for model inference. However, the computation cost and energy consumption may not be reduced with these approaches. Another approach is to train a thinner and deeper network directly, with highway~\cite{srivastava2015training} or residual connections~\cite{he2015deep} to overcome the optimization issue~\cite{lu2015small}. The resulting model is much more compact yet still accurate, e.g., it can achieve comparable recognition accuracy with around 10\% of the model parameters of a regular model on the AMI speech recognition corpus. 

The binary weight and activation approach in this paper differs from previous works in that it does not aim at cutting down the number of model parameters to save memory and computation, but to reduce the number of bits to save the weights and activations, and to turn the multiplications into additions and subtractions to save computation. Obviously, this approach is complementary to prior ones, and combinations with those approaches are possible, but they are not studied in this work. 

\section{Binary Neural Network}
\label{sec:bnn}
 
The key building block in neural networks is the linear matrix-vector multiplication followed by a nonlinear activation function such as
\begin{align}
\label{eq:nn-layer}
\hat{\bm h}_l &= \bm w_l \bm h_{l-1} + \bm r_l \\
\label{eq:func}
\bm h_l &= f_l(\hat{\bm h}_l),
\end{align}
where $\hat{\bm h}_l$ and $\bm h_l$ are activation vectors before and after the nonlinear function $f_l$; $\bm w_l, \bm r_l$ are the weight matrix and bias vector for the $l$-th layer. Most neural networks use real-valued weights $\bm w_l$ and activations $\bm h_l$ to preserve high precision. In this work, we explore the use of binary values for the weights $\bm w_l$ and activations $\bm h_l$ for acoustic modeling. 

\subsection{Binary Weights}
\label{sec:bin-wei}

In most of our work, we consider the binary pair $(-1, +1)$ instead of $(0, 1)$. While they are almost the same in terms of hardware implementation, neural networks with binary weights as $(-1, +1)$ may have a larger expressive power because of the subtraction operation corresponding to $-1$. In terms of model training, there are two ways to binarize the weight elements as discussed in~\cite{courbariaux2015binaryconnect}: stochastic and deterministic approaches. The stochastic approach is to set a weight $w$ to be $+1$ or $-1$ based on a probabilistic distribution, e.g., $p(\sigma(w))$, where $\sigma$ is the Sigmoid function that maps the real-valued $w$ to $[0, 1]$. The deterministic approach is to simply set $w$ to be its sign. As the deterministic approach is simpler for model training and inference, we focus on this approach in this paper. To be specific, the binarization function is the Sign function (also known as Hard Tanh function)
\begin{equation}
\label{eq:sign}
w^b = \text{Sign}(w) = \left \{ 
\begin{array}{l}
 +1 \quad \text{if  } w > 0 \\
 -1 \quad \text{otherwise} 
\end{array} \right .
\end{equation}
where $w^b$ is the binary weight and $w$ is the real-valued weight. Training the neural network with binary weights using stochastic gradient descent (SGD) is straightforward, as shown by Algorithm \ref{alg:bin-weight} following~\cite{courbariaux2015binaryconnect}. Note that we use $g_\theta$ to represent the gradient for $\theta$ as in the algorithm. Since we use $\hat{\bm h}_l$ to represent the activation vector before applying the nonlinear operation, the gradient through the nonlinear function can be written as $\frac{\partial f_l(\hat{\bm h}_l)}{\hat{\bm h}_l}$ (line 9). There are a few subtle points in this algorithm. Firstly, the gradients are not binary but are always real-valued (line 10 - 12). Secondly, the binary weights are only used for forward and backward propagations (line 3 and 10), and we always update the real-valued weights (line 17). The idea is to accumulate the gradient updates over multiple mini-batches. If we directly update the binary weight, then the gradient will make no effect in training if it is not large enough to flip the sign of the corresponding weight. For real-valued weights, however, the update will be accumulated, and the sign of the weight may be flipped after seeing a few more mini-batches. 

Since we use the binary weights for forward and backward propagations, while the real-valued weights are updated, there is an obvious mismatch between the gradient accumulation and model update. Such mismatch can cause optimization instability or even divergence. In order to narrow the gap, we applied two optimization tricks in our experiments.  The first one is clipping the real-valued weights after each update to prevent them from drifting far away from $\pm 1$ as used in~\cite{courbariaux2015binaryconnect}, i.e.,
\begin{equation}
\text{Clip}(w) = \left \{ 
\begin{array}{l}
+1 \quad \text{if } w> +1 \\
-1 \quad \text{if } w<-1 \\
w \quad \text{otherwise}
\end{array} \right . 
\end{equation} 
The second trick is that we optionally set each weight $w$ to its sign based on the probability $p(|w|)$ (line 19 - 21), where $|w|$ is the absolute value of $w$. Note that after clipping, $w \in [-1, +1]$. This is similar to the stochastic binarization approach as mentioned before. The idea is that when $|w_{ij}|$ is close to $1$, we set the weigh to its sign with a high probability to bridge the gap between the gradient accumulation and the model update. This weight will have less opportunity to change its sign, and can be considered to be locked for a few mini-batches. The weight that is closer to 0 will have a smaller probability to be set as its sign, so that it will still be active in training. This approach may have a similar effect to Dropout~\cite{srivastava2014dropout} to prevent coadaptation of the weights. Note that the process is stochastic, and a non-active weight may be active again after a few mini-batches. We refer to this approach as {\it semi-stochastic binarization} in this work. 

\begin{figure}[t]
\small
\centerline{\includegraphics[width=0.4\textwidth]{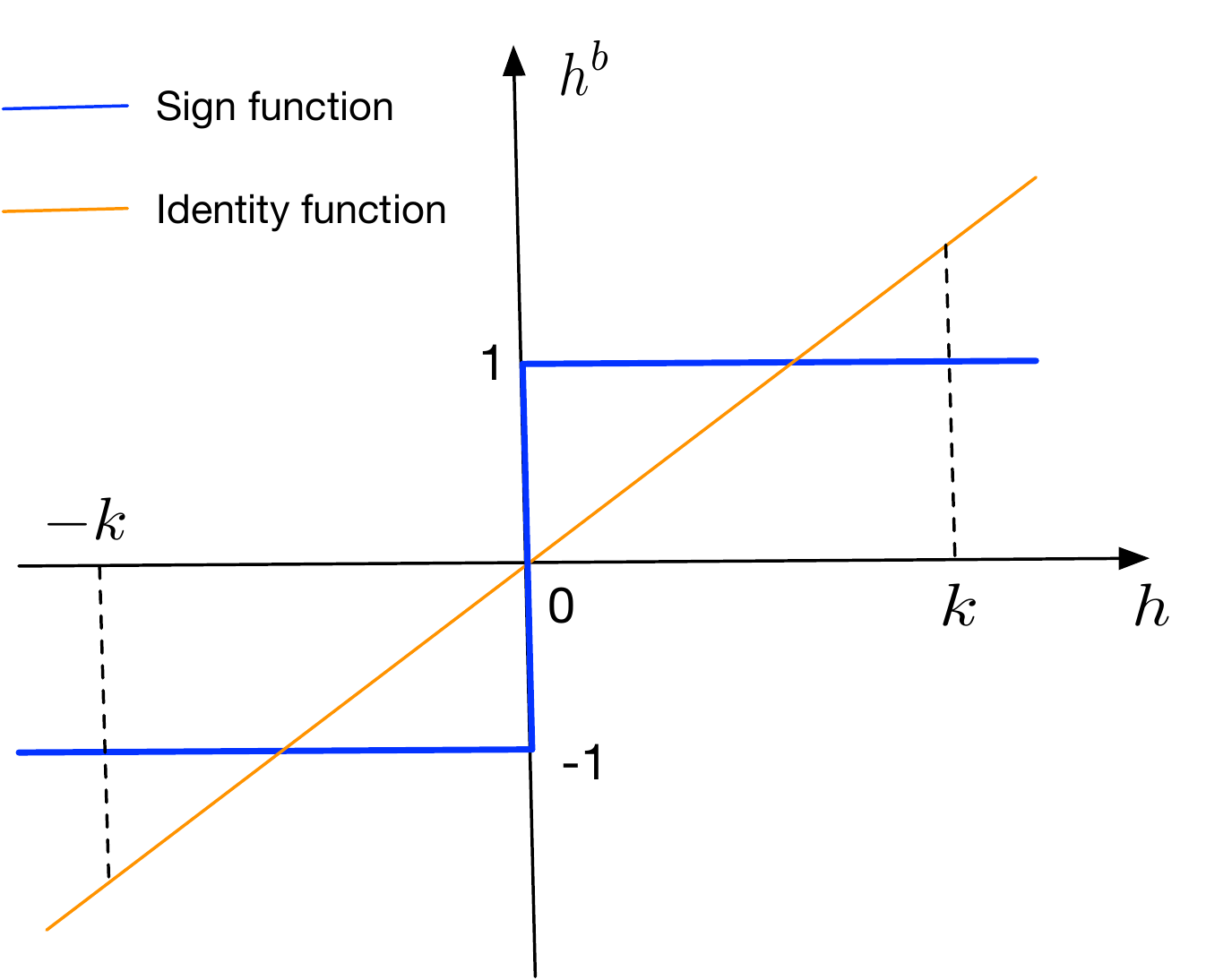}} \vskip -4mm
\caption{The blue line represents the Sign function, which is not differentiable. We use an Identity function (the orange line) to approximate it for back-propagation. When the input is beyond the range $[-k, k]$, the approximation error is considered to be large, and the corresponding gradient is set to be zero to disable the update. }  
\label{fig:bin-act}
\vskip -5mm
\end{figure}

\begin{algorithm}[t]                      
\caption{Forward-Backward propagation for feedforward neural networks with binary weights for all hidden layers.}          
\label{alg:bin-weight}                           
\begin{algorithmic}[1]                    
    \Require A minibatch of input and output samples, the loss of this minibatch $E$ and the learning rate $\eta$. $\bm h_0$ corresponds to the input feature vector.  
   \Function{Forward-Backward Propagation}{}
    \For{$l =1$ to $L$}  \Comment{Forward prop}
    \State $\bm w_l^b = \text{Sign}(\bm w_l)$ \Comment{Binarize the weight matrix}
    \State $\hat{\bm h}_l = \bm w_l^b \bm h_{l-1} + \bm r_l$
    \State $\bm h_l = f_l(\hat{\bm h}_l)$ \Comment{$f_l$: nonlinear function for $l$-th layer}
    \EndFor
    \State $g_{\bm h_L} =\frac{\partial E}{\partial \bm h_L}$ \Comment{Gradients from the loss function}
        \For{$l = L $ to $1$} \Comment{Backward prop}
    \State $g_{\hat{\bm h}_l}$ = $g_{\bm h_l}\frac{\partial f_l (\hat{\bm h}_l)}{\partial \hat{\bm h}_l}$ \Comment{Gradients through $f_l$}
    \State $g_{\bm h_{l-1}} = g_{\hat{\bm h}_{l}} \bm w_l^b$  
    \State $g_{\bm w_l^b} = g_{\hat{\bm h}_l} \bm h_{l-1} $
    \State $g_{\bm r_l} = g_{\hat{\bm h}_l} $
    \EndFor
   \EndFunction
   
    \Function{Update}{}
    \For{$l =1$ to $L$}
    \State $\bm w_l \leftarrow \text{Clip}(\bm w_l  - \eta g_{\bm w_l^b})$ \Comment{Update real-valued weights}
    \State $\bm r_l \leftarrow \bm r_l -\eta g_{\bm r_l}$ 
     \For{each element $w_{ij}$ in $\bm w_l$} \Comment{Optional }
      \State $w_{ij} = \text{Sign}(w_{ij}) \text{  with probability } p(|w_{ij}|)$
     \EndFor
    \EndFor
   \EndFunction
\end{algorithmic}
\end{algorithm}

\subsection{Binary Activations}
\label{sec:bin-act}

To binarize the activations, we use the same Sign function as Eq. \eqref{eq:sign} to binarize the activations. So in this case, the Sign function is the nonlinear function $f_l(\cdot)$ in Eq. \eqref{eq:func} for the binarization layer. Unlike the case of binary weights, however, using the Sign function as the nonlinear activation function will break the backpropagation algorithm as it is not differentiable. To address this problem, we use a function that is differentiable to approximate the Sign function during the backpropogation. In practice, we find that the {\it Identity} function works well, though other options may exist. The advantage of using the Identity function for the approximation is that it directly passes through the gradient without any change during backpropogation, which saves computation. However, this is an obviously biased estimate, and as Figure \ref{fig:bin-act} shows, the approximation error grows for larger input. In our experiments, training the neural network with binary activations did not converge by applying this approximation directly. To reduce the approximation error, we apply a mask to the gradient, so that when the absolute value of an element in $\hat {\bm h}_l$ is above a threshold $k$, where $k>0$, we set the corresponding gradient to be zero, meaning that the approximate error is unacceptable for that element, and the gradient is too noisy. In this case, the gradients through the binarization layer become
\begin{equation}
 g_{\hat{\bm h}_l} = g_{\bm h_l}\odot \text{Mask}(\hat{\bm h}_l, k)
\end{equation}
where $g_{\bm h_l}$ is the gradient before the binarization layer, and $g_{\hat{\bm h}_l}$ is the gradient after the binarization layer; $\odot$ is the elementwise multiplication. The mask is computed as 
\begin{equation}
\text{Mask}(h, k) = \left \{
\begin{array}{l}
0 \quad \text{if } |h| >k \\
1 \quad \text{otherwise}
\end{array} \right .
\end{equation}
In our experiments, only a small fraction of the hidden units in $\hat{\bm h}_l$ are above the threshold, so the gradient matrix $g_{\bm h_l}$ is very sparse. We may be able to take advantage of the sparsity of $g_{\bm h_l}$ to speed up the training, however, this was not investigated in this paper.  Courbariaux et al. \cite{courbariaux2016binarized} used the same approach for the approximated estimation, but the authors did not provide clear explanations for the motivation behind it, and $k$ was hard coded to $1$. The algorithm of forward and backward propagation with binary activations is summarized as Algorithm \ref{alg:bin-act}.

\begin{algorithm}[t]                      
\caption{Forward-Backward propagation for feedforward neural networks with binary activations for all hidden layers}          
\label{alg:bin-act}                           
\begin{algorithmic}[1]                    
    \Require A minibatch of input and output samples, the loss of this minibatch $E$, the learning rate $\eta$ and the threshold $k$. $\bm h_0$ corresponds to the input feature vector.  
   \Function{Forward-Backward Propagation}{}
    \For{$l =1$ to $L$}  \Comment{Forward prop}
    \State $\hat{\bm h}_l = \bm w_l \bm h_{l-1} + \bm r_l$
    \If {$l < L$}
    \State $\bm h_l = \text{Sign}(\hat{\bm h}_l$) \Comment{Binary activation function}
    \Else
    \State $\bm h_l = f_l(\hat{\bm h}_l)$ \Comment{Softmax function}
    \EndIf
    \EndFor
    \State $g_{\bm h_L} =\frac{\partial E}{\partial \bm h_L}$ \Comment{Gradients from the loss function}
        \For{$l = L $ to $1$} \Comment{Backward prop}
        \If{$l < L$}
          \State $g_{\hat{\bm h}_l} = g_{\bm h_l}\odot \text{Mask}(\hat{\bm h}_l, k)$ \Comment{Gradient through binarization layer}
        \Else
            \State $g_{\hat{\bm h}_l}$ = $g_{\bm h_l}\frac{\partial f_l (\hat{\bm h}_l)}{\partial \hat{\bm h}_l}$ \Comment{Gradient through Softmax layer}

        \EndIf
    \State $g_{\bm h_{l-1}} = g_{\hat{\bm h}_{l}} \bm w_l$  
    \State $g_{\bm w_l} = g_{\hat{\bm h}_l} \bm h_{l-1} $
    \State $g_{\bm r_l} = g_{\hat{\bm h}_l} $
    \EndFor
   \EndFunction
   
    \Function{Update}{}  \Comment{Standard parameter update}
    \For{$l =1$ to $L$}
    \State $\bm w_l \leftarrow \bm w_l  - \eta g_{\bm w_l}$ 
    \State $\bm r_l \leftarrow \bm r_l -\eta g_{\bm r_l}$ 
    \EndFor
   \EndFunction
\end{algorithmic}
\end{algorithm}

\subsection{Binary Neural Networks}
\label{sec:bin-nn}

Neural networks with both binary weights and activations are referred to as binary neural networks in this paper. Binary neural networks can further save computational cost as the matrix-vector multiplications become only {\tt XOR} operations in this case, and that can be quickly computed by hardware implementations. Training binary neural networks can be done by combining Algorithm \ref{alg:bin-weight} and \ref{alg:bin-act}. However, in our experiments, we observed that the gradients of the weight $g_{\bm w_l}$ can easily explode, resulting in divergence in training. We address the problem by clipping the norm of the gradients following the practice in training recurrent neural networks~\cite{pascanu2013difficulty} as
\begin{align}
  \text{if} ( \| g_{\bm w_l} \|_2  > \alpha) \quad \text{then} \quad g_{\bm w_l} \leftarrow \frac{\alpha}{\| g_{\bm w_l}\|_2 }  g_{\bm w_l}.
\end{align}
where $\| g_{\bm w_l}\|_2$ denotes the $\ell_2$ norm of $g_{\bm w_l}$, and $\alpha$ is the threshold. Note that a fully binary neural network is unlikely to work, and using binary weights for the Softmax layer is extremely harmful from our experience. In this work, binary neural networks only have binary weights and activations in the intermediate hidden layers.

\section{Experiments and Results}
\label{sec:results}

Most of our experiments were performed on the WSJ1 corpus, which has around 80 hours of training data, and we performed some additional experiments on the AMI meeting speech transcription task detailed in section \ref{sec:exp-ami}. In our experiments, we did not measure the computational cost, as the efficient computation with binary weights and activations relies on hardware implementations. Standard CUDA kernels for computations on GPUs do not have efficient ways to deal with multiplications with binary values. For the experiments on WSJ1, we used 40 dimensional log-mel filter banks with first oder delta coefficients as features, which were then spliced by a context of 11 frames. For acoustic modeling, we used feedforward neural networks with 6 hidden layers, with 3307 units in the Softmax layer, which is the number of the tied hidden Markov model triphone states. For the baseline models, we used Sigmoid activations for the hidden layers. Following the Kaldi recipe~\cite{povey2011kaldi}, we used the expanded dictionary and a trigram language model for decoding. All of our models were trained using SGD with exponential learning rate decay, and we used the cross entropy training criterion for all our systems. The algorithms for training neural networks with binary weights and activations are implemented within the Kaldi toolkit.

\begin{figure}[t]
\small
\centerline{\includegraphics[width=0.5\textwidth]{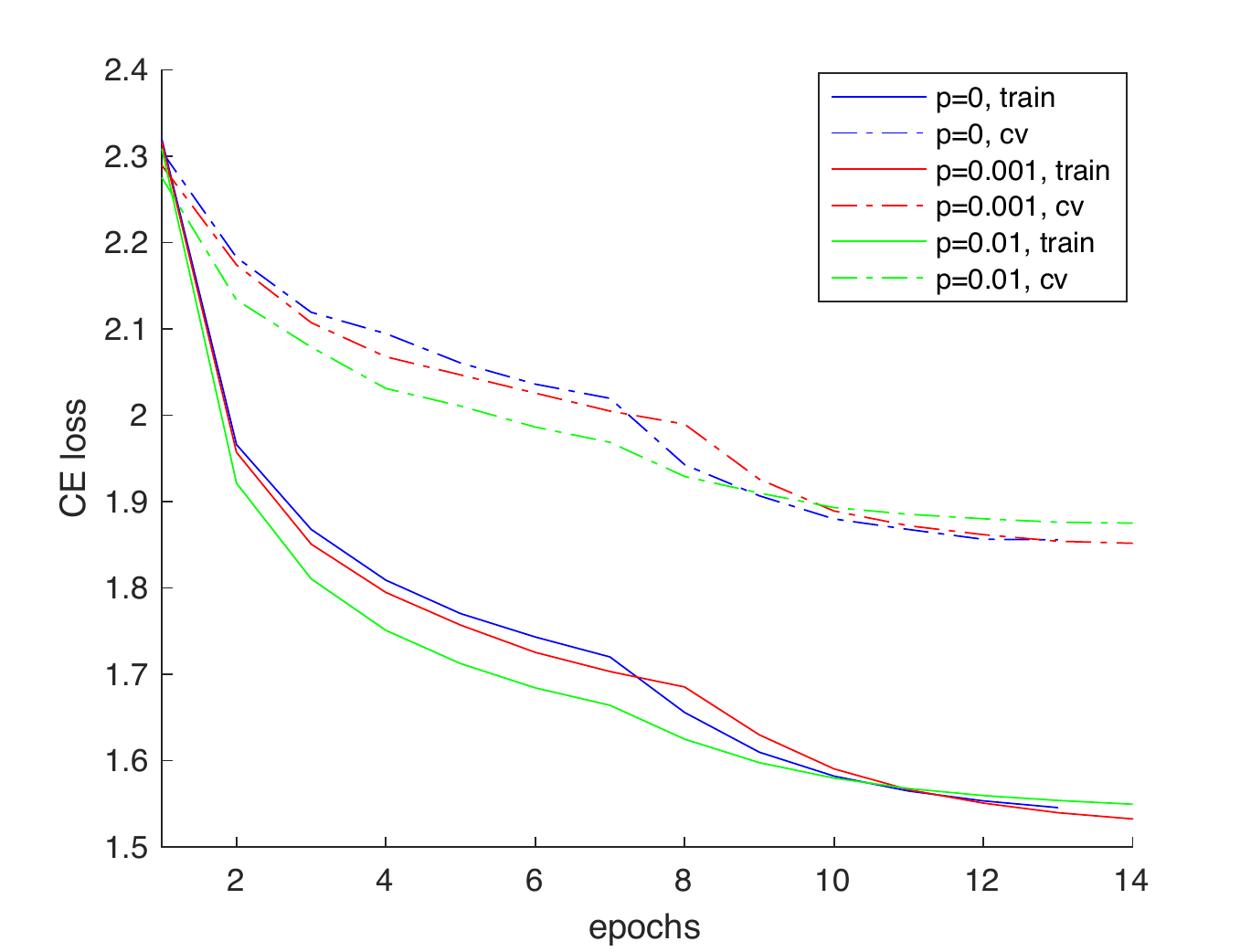}} \vskip -2mm
\caption{Convergence curves of boosted training for neural networks with binary weights on the WSJ1 dataset. $p=0$ corresponds to the system without boosted training.}  
\label{fig:boost}
\vskip -2mm
\end{figure}

\begin{table}[t]\centering
\caption{WERs (\%) of binary weight networks on WSJ1. The number of hidden units is 1024 for experiments in this table.}
\label{tab:bin-weight}
\footnotesize
\vskip0.15cm
\begin{tabular}{lcccc}
\hline 

\hline
Model       &  Input     &  Softmax       &  { \tt dev93}  & {\tt eval92} \\ \hline
Baseline  & --- & --- &  6.8 & 3.8 \\
Binary weights ($p=0$)     & fixed	  & fixed           & 7.7 & 4.8\\
Binary weights ($p=.001$)     & fixed	  & fixed           & 8.0 & 4.5\\
Binary weights ($p=.01$)     & fixed	  & fixed           & 8.0 & 4.4\\
Binary weights ($p=0$)   & real	  & real           & 10.4 & 6.7 \\
Binary weights ($p=0$)    & binary	  & fixed           & 12.0 & 7.3 \\
Binary weights ($p=0$)    & binary	  & binary           & 19.0 & 12.0 \\
\hline

\hline
\end{tabular}
\vskip-3mm
\end{table}

\begin{figure}[t]
\small
\centerline{\includegraphics[width=0.5\textwidth]{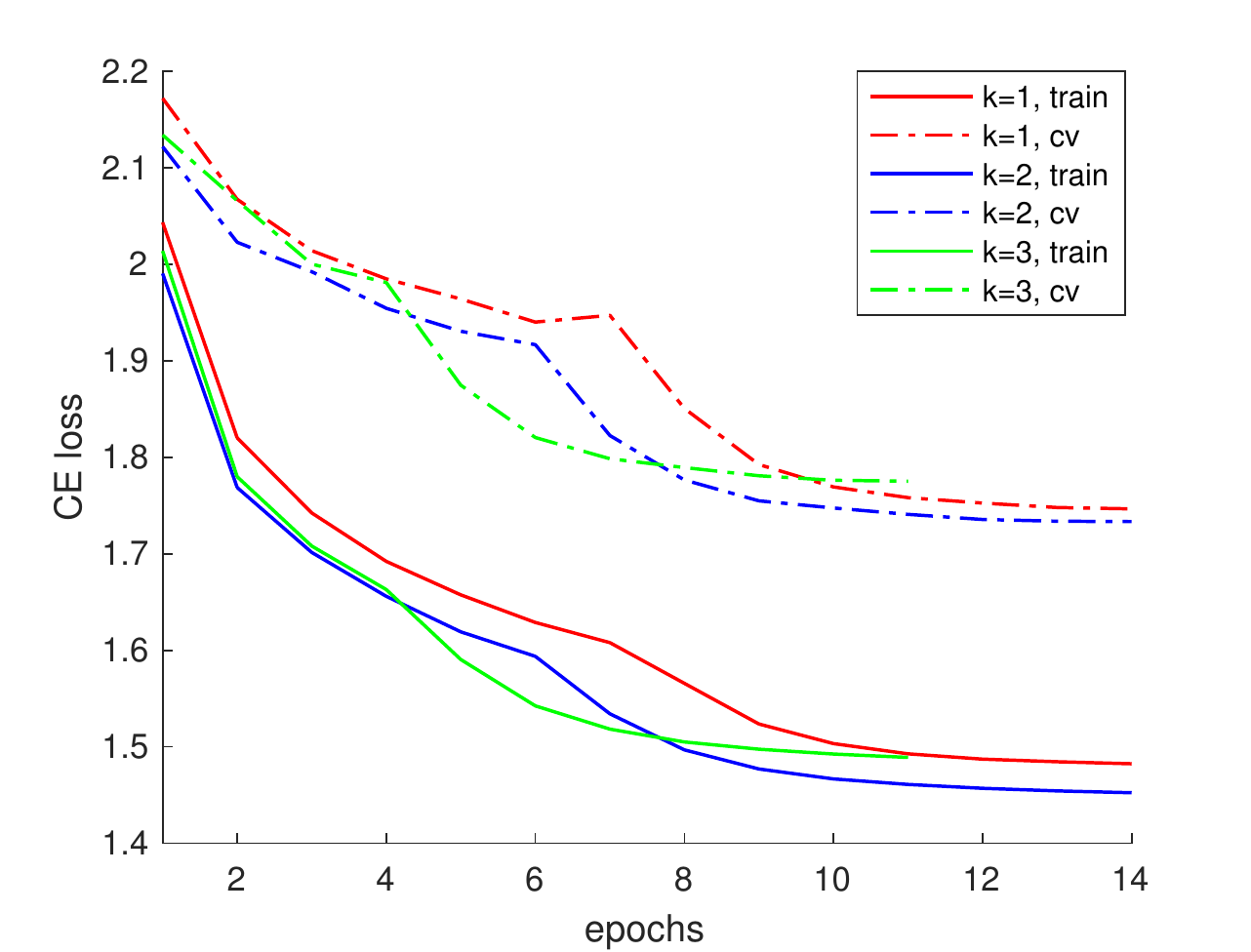}} \vskip -2mm
\caption{Convergence curves of neural networks with binary activations on the WSJ1 dataset. }  
\label{fig:value-k}
\vskip -2mm
\end{figure}

\begin{table}\centering
\caption{WERs (\%) of networks using binary activations. $\sigma$ denotes the Sigmoid activation, $b$ means binary activation, and $\delta$ represents the Softmax operation. }
\label{tab:bin-act}
\footnotesize
\vskip0.15cm
\begin{tabular}{ccccccc}
\hline 

\hline
        & & &          \multicolumn{2}{c}{Update layer}    &  \multicolumn{2}{c}{WER} \\ 
ID & Model        & $k$ &  Input      &  Softmax        &  {\tt dev93}  & {\tt eval92} \\ \hline
1 & ($\sigma, b, b, b, b, \sigma, \delta$)     & 1 & $\times$	  & $\times$           & 8.2 & 4.4 \\
2 & ($\sigma, b, b, b, b, \sigma, \delta$)     & 2 & $\times$	  & $\times$         & 7.8 & 4.8 \\
3 & ($\sigma, b, b, b, b, \sigma, \delta$)     & 3 & $\times$	  & $\times$          & 8.0 & 4.5 \\
4 & ($\sigma, b, b, b, b, \sigma, \delta$)     & 1 & $\checkmark$	  & $\checkmark$           & 9.1  & 5.3 \\
5 & ($\sigma, b, b, b, b, \sigma, \delta$)     & 1 & $\checkmark$	  & $\times$          & 9.6 & 5.8 \\ \hline
6 & ($b, b, b, b, b, \sigma, \delta$)     & 1 & $\times$	  & $\times$           & 8.1 & 4.8 \\ 
7 & ($b, b, b, b, b, \sigma, \delta$)     & 1 & $\checkmark$	  & $\times$           & 11.3 & 7.0 \\ 
8 & ($b, b, b, b, b, \sigma, \delta$)     & 1 & $\checkmark$	  & $\checkmark$           & 10.7  & 6.7 \\ \hline
9 & ($b, b, b, b, b, b, \delta$)     & 1 &$\times$	  & $\times$           & 20.4 & 14.5 \\
10 & ($b, b, b, b, b, b, \delta$)     & 1 & $\times$	  & $\checkmark$          & 12.1 & 7.0 \\
11 & ($b, b, b, b, b, b, \delta$)     & 1 & $\checkmark$	  & $\checkmark$          & 20.5 & 12.7 \\
\hline

\hline
\end{tabular}
\vskip-3mm
\end{table}

\subsection{Results with binary weights}

In our experiments, training neural networks with binary weights from random initialization usually did not converge, or converged to very poor models. We addressed this problem by initializing our models from a well-trained neural network model with real-valued weights. This approach worked well, and was used in all our experiments. We trained the baseline neural network with an initial learning rate of 0.008, following the {\tt nnet1} recipe in Kaldi. We then reduced the initial learning rate to 0.001 when running Algorithm \ref{alg:bin-weight} to train the neural network with binary weights. This initial learning rate was found to be a good tradeoff between convergence speed and model accuracy. Table \ref{tab:bin-weight} shows the word error rates (WERs). Here, we explored several settings: the weights in the input layer and Softmax layer were binary, real-valued or fixed from initialization. As shown by Table \ref{tab:bin-weight}, fixing the weights of the input and Softmax layers to be initialized values and only updating the binary weights of the intermediate hidden layers achieves the best results, which are around 1\% absolute worse than our baseline. We also did experiment to update those real-valued weights jointly with the binary weights using the same learning rate, but obtained much worse results. The reason may be that the gradients of the real-valued weights and binary weights are in very different ranges, and updating them using the same learning rate is not appropriate. Adaptive learning rate approaches such as Adam~\cite{kingma2014adam} and Adagrad~\cite{duchi2011adaptive} may work better in this case, but they are not investigated this work. In order to have a complete picture, we have also tried using binary weights for the input layer and the Softmax layer. As expected, we achieved much lower accuracy, confirming that reducing the resolution of the input features and activations for the Softmax classifier are harmful for classification.

We then studied the semi-stochastic binarization approach in Algorithm \ref{alg:bin-weight} (line 19 - 21). Applying this step very frequently is harmful to the SGD optimization as it can counteract the SGD update. In our experiments, we set a probability $p$ to control the frequency of this operation. More precisely, after each SGD update, we draw a sample from a uniform distribution between 0 and 1, and if its value is smaller than $p$, then the semi-stochastic binarization approach will be applied. Therefore, a larger $p$ means more frequent operations and vice versa. Figure \ref{fig:boost} shows the convergence curves of training with and without this approach, suggesting that the semi-stochastic binarization can speed up the convergence. However, as Table \ref{tab:bin-weight} shows, we did not achieve consistent improvements on the {\tt dev93} and {\tt eval92} sets. Note that, we used the {\tt dev93} set to choose the language model score for the {\tt eval92} set, and from our observations, the results of development and evaluation sets are usually not well aligned, demonstrating that there may be a slight mismatch between the two evaluation sets. The semi-stochastic binarization approach will be revisited on the AMI dataset.

\begin{table}\centering
\caption{WERs (\%) of neural networks with binary weights and activations on the WSJ1 dataset. We set $p=0$ for the system with binary weights, and $k=1$ for the system with binary activations.}
\label{tab:bnn}
\footnotesize
\vskip0.15cm
\begin{tabular}{lcccc}
\hline 

\hline
Model       &  Size        & $b$   &  {\tt dev93}  & {\tt eval92} \\ \hline
Baseline & 1024 & -- &  6.8 & 3.8 \\
Baseline & 2048 & -- & 6.5 & 3.5 \\
Binary weights & 1024  & $(-1, +1)$ & 7.7 & 4.8 \\
Binary weights & 1024  & $(0, 1)$ &  \multicolumn{2}{c}{Not Converged}  \\
Binary activations  & 1024 & $(-1, +1)$ & 8.2 & 4.4 \\
Binary activations  & 1024 & $(0, 1)$ & 7.2 & 4.1 \\
Binary neural network & 1024 & $(-1, +1)$ &  15.6 & 10.7 \\ \hline
Binary activations  & 2048  & $(-1, +1)$  & 7.3 & 4.4 \\
Binary weights  & 2048 & $(-1, +1)$   & 7.5 & 4.4 \\
Binary neural network & 2048 & $(-1, +1)$  & \multicolumn{2}{c}{Not Converged}  \\

\hline

\hline
\end{tabular}
\vskip-3mm
\end{table}

\subsection{Results with binary activations}

Following the previous experiments, we also initialized the model from the well-trained real-valued neural network for the binary activation networks. Again, we set the initial learning rate for the binary activation network to 0.001. In the first set of experiments, the activation functions of the first and last hidden layers of the network were fixed to Sigmoid activations, and only those of the hidden layers in between were replaced by binary activations. We first studied the impact of the hyper-parameter $k$ in our model training. As mentioned before, smaller $k$ corresponds to more sparse gradients, while larger $k$ indicates a larger approximation error. From our experiments, setting $k$ between 1 and 3 did not make a big difference in terms of WERs for both evaluation sets. Figure \ref{fig:value-k} shows that $k=2$ is a good tradeoff between convergence speed and model generalization ability. With $k=4$, however, the model training did not converge in our experiments due to the large approximation error.

We also looked at updating the weights in the input and Softmax layers in this case. As the table shows, keeping both layers fixed still works the best. Again, this may be due to the fact that the gradients from Sigmoid and binary activations are in different ranges. In the future, we shall revisit this problem with adaptive learning rate approaches. We then investigated using binary activations for the first hidden layer (row 6 - 8) and the last hidden layer (row 9 - 11).  Surprisingly, when the weights in both input layer and Softmax layer are fixed, using binary activations for the first hidden layer can achieve comparable accuracy in our experiments. However, using binary activations for the last hidden layer degraded the accuracy remarkably, which is expected as the resolution of the features for the Softmax layer is very low in this case. 

Table \ref{tab:bnn} shows results of networks with a larger number of hidden units and binary neural networks with both binary weights and activations. For all the experiments in this table, the weights in the input and Softmax layer were fixed, and the first and last hidden layers used Sigmoid activations. Using a larger number of hidden units works slightly better for both binary weight and binary activation systems. For the binary neural network system, we applied the gradient clipping approach as explained in section \ref{sec:bnn} to prevent divergence in training, and set $\alpha$ to 15. However, we only managed to train the network with 1024 hidden units and achieved much inferior accuracies. Training fully binary neural networks is still a challenge from our study. 

We also did some experiments to compare $(-1, +1)$ to $(0, 1)$ for binarization as shown in Table \ref{tab:bnn}. With Sigmoid activations, using (0, 1) for binary weights can cause training divergence as the elements of the activation vector $\hat{\bm h}_l$ are always positive. Using $(0,1)$ for binary activations, however, we achieved lower WER. The reason may be that the network was initialized from Sigmoid activations, and $(0, 1)$ is much closer to Sigmoid compared to $(-1, 1)$. In fact, $(0,1)$ binary function can be viewed as the hard version of Sigmoid. Using $(-1, +1)$ for binary activations may work better with networks initialized from Tanh activations, and that will be investigated in our future works.  







\begin{table}[t]\centering
\caption{WERs (\%) of neural network with binary weights and activations on the AMI dataset. The number of hidden units is 2048, and $b$ denotes binarization.}
\label{tab:ami}
\small
\vskip0.15cm
\begin{tabular}{lccc}
\hline 

\hline
Model       &  $b$           &  {\tt dev}  & {\tt eval} \\ \hline
Baseline & -- &  26.1 & 27.5 \\
Binary weights ($p = 0$) & $(-1, +1)$  & 30.3 & 32.7 \\
Binary weights ($p = .001$) & $(-1, +1)$  & 30.0 & 32.2 \\
Binary weights ($p = .01$) & $(-1, +1)$  & 29.6 & 31.7 \\
Binary weights ($p = .05$) & $(-1, +1)$  & 29.6 & 31.9 \\
Binary activations ($k=1$)  & $(-1, +1)$  & 30.1 & 32.5 \\
Binary activations ($k=2$)  & $(-1, +1)$  & 29.9 & 32.3 \\
Binary activations ($k=3$)  & $(-1, +1)$  & 30.2 & 32.4 \\
Binary activations ($k=4$)  & $(-1, +1)$  & 29.8 & 32.0 \\
Binary activations ($k=1$)  & $(0, 1)$  & 27.5 & 29.5 \\
Binary activations ($k=2$)  & $(0, 1)$  & 28.0 & 30.2 \\
Binary activations ($k=3$)  & $(0, 1)$  & 29.8 & 32.2 \\
Binary neural network & $(-1, +1)$ &   \multicolumn{2}{c}{Not Converged}  \\ 

\hline

\hline
\end{tabular}
\vskip-3mm
\end{table}

\subsection{Results on the AMI dataset}
\label{sec:exp-ami}

As we mentioned before, we did not observe consistent trends on the development and the evaluation sets of WSJ1, possibly due to certain mismatch between the two. This hindered us from drawing strong conclusions. To gain further insights on the techniques that we have explored, we performed some experiments on the AMI corpus. We focused on the IHM (individual head microphone) condition. It also has around 80 hours of training data, but the {\tt dev} and {\tt eval} sets are much larger (over 8 hours). Again, we built our baseline following the Kaldi recipe. We used MFCC features followed by feature-space MLLR transformation, and a trigram language model for decoding. The neural network models have 6 hidden layers, and the Softmax layer has 3972 units. As in the WSJ1 experiments, we initialized our model from the baseline model for networks with binary weights and binary activations. The initial learning rate is 0.008 for the baseline system, and 0.001 for binary weight and activation systems. 

The experimental results are shown in Table \ref{tab:ami}. For the binary weight systems, we revisited the semi-stochastic binarization approach. While the convergence curves were similar to Figure \ref{fig:boost} in this case (not shown in this paper), we obtained small but consistent improvements on both of {\tt dev} and {\tt eval} sets. In particular, with $p=0.01$, the improvement is around 1\% absolute on the {\tt eval} set as shown by Table \ref{tab:ami}. Since the model was initialized from a Sigmoid network, the binary activation system with $(0,1)$ worked much better than its counterpart with $(-1,+1)$ when $k=1$. Again, we did experiments by tuning the threshold $k$ for binary activation systems. Unlike the experiments on WSJ1, the models are relatively tolerant to changing $k$ with $(-1, +1)$ for binarization, and we only observed divergence when $k=7$. This may be because that we used different features in these experiments, causing differences in the distributions of $\hat{\bm h}_l$. However, this is not the case for binarization with $(0,1)$, as the system degrades  rapidly when $k$ increases. The reason may be that the binarization function with $(0,1)$ is not symmetric, and the approximation error using an identity function is significant for large $k$ when inputs are negative.  Again, we failed to train binary neural networks with 2048 hidden units due to divergence in training.

\section{Conclusion}

Neural networks with binary weights and activations are appealing for deploying deep learning applications on embedded devices. In this paper, we investigated this kind of neural networks for acoustic modeling. In particular, we have presented practical algorithms to training neural networks with binary weights and activations, and discussed optimization techniques to handle training divergence. On both WSJ1 and AMI datasets, we have achieved encouraging recognition WERs compared to the baseline models. However, this study is still in the early stage, and there is still much room to explore. For example, we only considered feedforward neural networks in this work, leaving other neural architectures such as convolutional neural network and recurrent neural networks, as open problems. Training the network with both binary weights and activations is still challenging from our results, and more work is needed to address the optimization challenge for this case.

\section{Acknowledgements}

We thank the NVIDIA Corporation for the donation of a Titan X GPU used in this work, and Karen Livescu for proofreading and comments that have improved the manuscript. 

\bibliographystyle{IEEEtran}
\bibliography{bibtex}

\end{document}